\pdfoutput=1

\documentclass[11pt]{article}

\usepackage{acl}

\usepackage{times}
\usepackage{latexsym}

\usepackage[T1]{fontenc}

\usepackage[utf8]{inputenc}

\usepackage{microtype}

\usepackage{inconsolata}

\usepackage{hhline}
\usepackage{graphicx}
\usepackage[export]{adjustbox}
\usepackage{multibib}
\usepackage{tabularx}
\usepackage{epstopdf}
\usepackage{arabtex}
\usepackage{utf8}   
\usepackage{hyperref}
\usepackage{xstring}
\usepackage{color}
\usepackage{multirow}
\usepackage{soul}
\usepackage{titlesec}
\usepackage{amssymb}
\usepackage{colortbl}
\usepackage{makecell}

\setcode{utf8}
\usepackage{abstract}
\title{A Benchmark and Scoring Algorithm for Enriching Arabic Synonyms}

\author{Sana Ghanem$^{1}$, Mustafa Jarrar$^{1}$, Radi Jarrar$^{1}$, Ibrahim Bounhas$^{2, 3}$ \\
  $^{1}$Department of Computer Science, Birzeit University,  Palestine \\
  $^{2}$LISI Laboratory of Computer Science for Industrial System, INSAT, Carthage University, Tunisia \\
  $^{3}$JARIR: Joint group for Artificial Reasoning and Information Retrieval, Tunisia \\
  \texttt{\{swghanem, mjarrar, rjarrar\}@birzeit.edu} \\
  \texttt{ibrahim.bounhas@isd.uma.tn}
 }

\begin{document}
\maketitle
\begin{abstract}

This paper addresses the task of extending a given synset with additional synonyms taking into account synonymy strength as a fuzzy value. Given a mono/multilingual synset and a threshold (a fuzzy value $[0-1]$), our goal is to extract new synonyms above this threshold from existing lexicons. We present twofold contributions: an algorithm and a benchmark dataset. The dataset consists of 3K candidate synonyms for 500 synsets. Each candidate synonym is annotated with a fuzzy value by four linguists. The dataset is important for (i) understanding how much linguists (dis/)agree on synonymy, in addition to (ii) using the dataset as a baseline to evaluate our algorithm. Our proposed algorithm extracts synonyms from existing lexicons and computes a fuzzy value for each candidate. Our evaluations show that the algorithm behaves like a linguist and its fuzzy values are close to those proposed by linguists (using RMSE and MAE). The dataset and a demo page are publicly available at \url{https://portal.sina.birzeit.edu/synonyms}.

\end{abstract} 

\section{Introduction and Motivation}
Synonymy relationships are used in many NLP tasks and knowledge organization systems. However, automatic synonym extraction is a challenging task, especially for low-resourced and highly ambiguous languages such as Arabic \cite{DH21}. There are some Arabic resources representing synonymy, such as Al-Maknaz Al-Kabīr, Arabic WordNet \cite{ArabicWordNet} and the Arabic Ontology \cite{J21,J11}; however, these resources are limited in terms of size and coverage \cite{APJ16,HJ21b}, especially if compared with the English Princeton WordNet \cite{miller1990introduction}. Building such resources is expensive and challenging \cite{HPJF14,JA19,J20}. In addition, the notion of synonymy itself is problematic, as it can vary from near (i.e., semantically related) to strict synonymy \cite{JKKS21,J05}. Strict and formal synonymy is used in ontology engineering as an equivalence relation, thus its reflexive, symmetric, and transitive \cite{J21}. A less formal synonymy is used in the construction of synsets in Princeton WordNet, which relies on the substitutionability of words in a sentence: ``two expressions are synonymous in a linguistic context \textit{c} if the substitution of one for the other in \textit{c} does not alter the truth value'' \cite{miller1990introduction}. For example, ({\scriptsize \<طريق>}/road) and ({\scriptsize \<شارع>}/street) are substitutionable in many contexts in Arabic, thus they can be synonyms. As will be reviewed in section \ref{sec:relatedword}, different approaches have been proposed for extracting synonyms automatically.

Nevertheless, one of the major challenges in extracting synonyms is that it is hard to evaluate them \citep{wu2003optimizing} and there are no common evaluation datasets. Moreover, the substitutionability criteria are subjective, because humans do not necessarily agree on synonymy. As will be illustrated later in this paper, if different linguists are given the same words to judge whether they are synonyms, it is unlikely that they will agree on all cases. Thus, instead of relying on ``the substitutionability of words in a sentence'' as a criterion to judge whether two words are synonyms or not, we propose to model it with a \textit{fuzzy value}. For example, let \{confederacy, confederation\} be two synonyms in the context of ``a union of political organizations'', and let ``alliance'' and ``federation'' be candidate additional synonyms, our goal is to assign a fuzzy value (e.g., 0.6 and 0.9) to each candidate synonym to indicate how much it is substitutionable, i.e., acceptable to be an additional third synonym. 

Using such a fuzzy value is helpful for different application scenarios. For example, when constructing wordnet synsets, synonyms can be extracted with a high fuzzy value, but in the case of less sensitive information retrieval applications, a lower value might be more suitable. In a quality control scenario, one may evaluate a thesaurus by masking each synonym in a synset and assessing if its fuzzy value passes a threshold. Nevertheless, assigning a meaningful fuzzy value to each synonym in a synset is challenging. Thesauri are typically constructed based on linguists' intuition and without assigning a strength, or a fuzzy, value explicitly. To overcome this challenge, we developed a dataset of 3K synonyms, each assigned with a fuzzy value by four different linguists. We used this dataset to measure how much linguists (dis/)agree on synonymy. The dataset is also used to train our proposed algorithm (i.e., tune its fuzzy model) for extracting synonyms from dictionaries.

\textbf{\textit{Task definition}}: The task we aim to address is defined as the following: Let $S$ be a set of synonyms, $c$ is a candidate synonym to S, and a dictionary $D$, our goal is to compute a fuzzy value $f$ to indicate how much $c$ is acceptable to be an addition to $S$. As will be elaborated in section \ref{sec:algorithm}, we assume $D$ to be a set of sets of synonyms, and that $S$ can be mono or multilingual synonyms.

Our main contributions in this paper are a dataset and an algorithm. The dataset was constructed by employing four linguists and giving them 3,000 candidate synonyms and 500 synsets from the Arabic WordNet \cite{ArabicWordNet}. Each linguist was asked to score each candidate synonym in a given synset. 
Our proposed algorithm aims at discovering new candidate synonyms from existing linguistic resources. Given a set of synonyms, the algorithm builds a directed graph, at level \textit{k} for all words in this set. Cyclic paths in this graph are then detected, and all words participating in these cyclic paths are considered candidate synonyms for the given synset. Each of these candidate synonyms is assigned a fuzzy value, which is calculated based on a fuzzy model that we learned from the dataset and that takes into account the connectivity of the candidate synonym in the graph.
The novelty of our algorithm and our dataset is that we treat synonyms as a fuzzy relation. We evaluated the algorithm's fuzzy values by comparing them with the average of the linguists' scores (i.e., as a baseline). The Root Mean Squared Error (RMSE) between the scores of the algorithm and the average of the linguists' scores is 0.32 and the Mean Average Error (MAE) is 0.27. This means that the algorithm was behaving closely to a linguist. To evaluate the accuracy of our algorithm, we used the 10K synsets in Arabic WordNet. We masked the word with (highest, lowest, average, and random) frequency in each synset and used the algorithm to see if it could discover it again with top rank. The achieved accuracy was indeed high. For example, with the average frequency we achieved an accuracy of 98.7\% at level 3 and 92\% at level 4. 

This paper is organized as follows: Section 2 presents related works in the field of synonym extraction. Section 3 overviews the algorithm. Section 4 summarizes and discusses the experimental results. Section 5 concludes the paper and proposes some perspectives.

\section{Related Work}
\label{sec:relatedword}

In what follows, we overview several approaches have been proposed to extract synonyms or build synsets.  We refer to \cite{KAJ21} for a recent survey on this topic.

\subsection{Synset Construction}
New WordNets may be built by mining corpora and/or monolingual dictionaries as in \cite{Oliveira14} for Portuguese. After extracting candidate synonym pairs, authors cluster these pairs into different clusters. 
\citet{ercan2019synset} proposed to build a multilingual synonymy graph from existing resources and wordnets, then used a supervised clustering algorithm to cluster synonyms. In both works, each cluster is then considered a synset. Neural language models, such as word embeddings, were also employed in synonymy extraction and wordnet construction \cite{Mohammed2020}. For example, \citet{khodak2017automated} proposed to construct wordnets using the Princeton WordNet (PWN), machine translation, and word embeddings. A word is first translated into English using machine translation, and these translations are used to build a set of candidate synsets from PWN. A similarity score is used to rank each candidate synset, which is calculated using the word embedding-based method. 
Similarly, \citet{tarouti16} used static word embeddings to improve the quality of automatically constructed Arabic wordnet. 
Furthermore, \citet{al2021synoextractor} proposed to extract Arabic synonyms based on a static word embedding model that was created using Arabic corpora. Cosine similarity, in addition to some filters, were used to extract Synonyms.

\subsection{Synonym Graph Mining}
Other approaches are proposed to mine a graph from an existing resource(s) in order to discover new synonyms and translation pairs. The structure of the graph is exploited to compute ranking scores, which reflect how much two terms are likely to be synonyms \cite{J05}. The main hypothesis is that some words, which are not necessarily directly connected with an edge may be semantically close. That is why cycles are widely exploited. Indeed, graphs are generic tools that may be used both for monolingual and bilingual resources and for several types of linguistic resources. 
For example, \citet{flati2012cqc} proposed an algorithm to find missing synonyms in the Ragazzini-Biagi English-Italian dictionary. A synonymy graph was built using this dictionary, then cyclic and quasi-cyclic paths are detected. Cyclic paths are those that have all edges in the same direction, while quasi cycles should be consecutive reverse edges. The length of a path is used to score the discovered synonyms. Discovering new translation pairs from multilingual dictionaries is also related to synonymy extraction. \citet{villegas2016leveraging} proposed to construct a multilingual translation graph using translation pairs in the Apertium dictionaries. New translation pairs are then extracted from cyclic paths. However, wrong translations might be detected because of polysemy. The authors proposed to score the density of each path and exclude those paths with low densities. Instead of only using density, \citet{torregrosa2019tiad} proposed to combine it with a multi-way neural machine translation trained with parallel English and Spanish, Italian and Portuguese, and French and Romanian corpora. Their experiment shows a low recall and a reasonable precision ($25\%-75\%$). 

A recent algorithm that uses synonymy graphs was proposed by \citet{JKKS21}. The idea is to construct an Arabic-English translation graph from a given bilingual dictionary \cite{JAM19}. Terms participating in cyclic paths are extracted and consolidated, and considered synonyms. However, instead of using fuzzy values, they proposed the idea of bidirectional consolidation.

\subsection{Related Notions of Fuzziness}
Different notions of fuzziness were proposed in the WordNet literature. 
\citet{iran2020} and \citet{iran2021} proposed to compute the frequency of each word-sense pair in a corpus that is annotated using a WSD algorithm. The frequency is then normalized and transformed into a “possibility” value between 0 and 1 reflecting the membership degree. 
In \cite{iran2020}, the same notion is evaluated in an interval indicating minimum and maximum values by dividing the corpus into several categories.  In both cases, 
These membership degrees depend on the number of times a word-sense pair appeared in a given corpus. 
We believe that this notion of fuzziness is valuable and complements our proposed work; however, it highly depends on the coverage of the used corpora and the accuracy of the WSD algorithm, which is typically not good enough \cite{nibbling2022}. Another notion of fuzziness was used in \cite{Oliveira2016}, to compute how likely two words are synonyms based on much they share words in their dictionary definitions. This notion of fuzziness was used to extract a Portuguese synonym network from seven resources taking into account the number of times a relation between two given words exists across resources. This notion of fuzziness, similar to \cite{iran2020}, depends on text mining rather than synonyms graphs. Additionally, it computes the fuzziness between two words rather than between a word and a given synset. Most importantly, as discussed in section \ref{sec:Scoring}, our fuzzy scores are designed to reflect meaningful values, i.e., semantic truth, rather than frequency of use.

\subsection{Benchmarks} As far as benchmarking and evaluation are concerned, it is hard to compare previous works, given the lack of a common gold standard. Indeed, the above-reviewed approaches were evaluated using different ways and resources, as no evaluation benchmarks are available for synonymy extraction. More precisely, and to our knowledge, there are no datasets of synonyms with ranking or fuzzy values to indicate how much a term is likely to be a synonym with a given synset.

\section{Dataset Construction}
\label{sec:dataset}
This section presents a benchmarking dataset annotated with fuzzy values\footnote{The dataset and source code are publicly available at \url{https://portal.sina.birzeit.edu/synonyms}}.
The dataset can be used for training and evaluating (i.e., a baseline) synonym extraction algorithms. Additionally, the construction of this dataset can also be used as an experiment to measure how much linguists (dis/)agree on synonymy.

\subsection{Data Selection}
First, we selected 500 synsets from the 10K synsets in Arabic WordNet. For each synset, we extracted a set of Arabic candidate synonyms, which we collected using our algorithm presented in Section \ref{sec:algorithm}. The total number of candidate synonyms is 3K. The 500 synsets were selected proportionally to the WordNet's distribution: 350 noun synsets, 140 verb synsets, and 10 adjective synsets. These synsets were selected randomly but we also took into account synset length and selected 142, 207, and 151 synsets of 2, 4, and 6 words in each synset, respectively. The 3K candidate synonyms were then given to four linguists to give them scores.

\subsection{Experimental Setup}
\begin{figure}[t]
    \centering
    \includegraphics[width=0.5\textwidth]{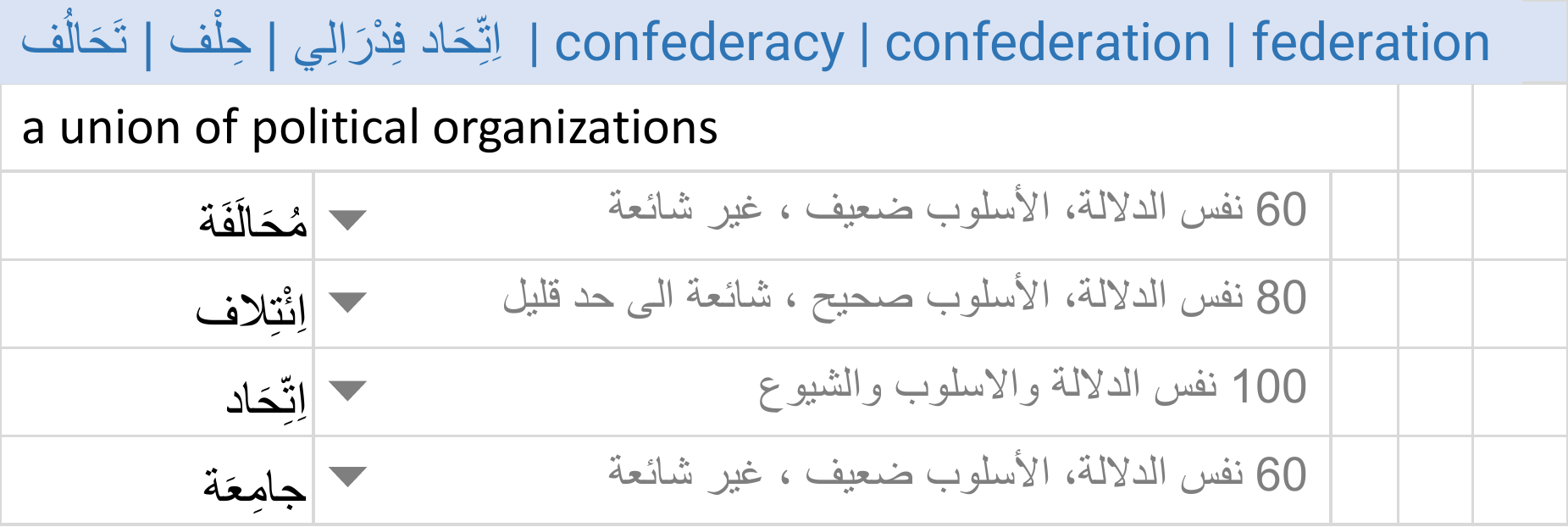}
    \caption{Example of scoring candidate synonyms.}
    \label{fig:ExampleGSheet}
\end{figure}

The four linguists who participated in this experiment are top students, who graduated recently with high distinction from the department of linguistics and translation at Birzeit University. Three training workshops were organized to explain the experiment and to emphasize the notion of synonymy. To ensure that all linguists have the same understanding of the task, we gave each linguist a small quiz ($\mathtt{\sim}$30 synonyms) to try alone, then we discussed the results jointly. After that, each linguist was given the 3K candidate synonyms in a separate file in Google Sheet. Figure~\ref{fig:ExampleGSheet} illustrates an example of a synset and four candidate synonyms as scored by one of the linguists. As shown in Figure~\ref{fig:ExampleGSheet}, the scoring is based on the linguist's understanding of the given synset (both English and Arabic synonyms), the gloss, and the context example (if available), which we extracted from the Arabic WordNet.

\subsection{Scoring Guidelines}
\label{sec:Scoring}
Table~\ref{table:SemanticMeanings} presents our  scoring schema, which is a scale from 0 to 100 representing the strength of the synonymy relation. The main factor in the scoring is the semantics, which indicates \textit{how much the truth of a sentence is altered if the candidate synonymy is substituted with one of the given synonyms}, as defined in \citet{miller1990introduction}. 
The scoring schema should not be interpreted as absolute numbers, but rather, they are used as annotation methodology to maintain a degree of consistency among linguists' scores as will be discussed next. From a semantics viewpoint, the scoring schema is divided into three categories: same $(>60\%)$, close $(60\%-50\%)$, or related/different semantics $(<50\%)$. \textit{Same semantics} means that a word can be substituted in a sentence without altering the truth of this sentence. The four different scores inside this range are used to capture the \textit{use}; i.e., how much it is common that a word can be used in this context. For example, the word {\scriptsize \<اِئْتِلاف>}
has the same semantics as the other synonyms in the synset that means ``a union of political organizations'', but this word is rarely used in this context. \textit{Close Semantics} means that it is possible to use a word (e.g., {\scriptsize \<جامِعَة>}) with this semantics, but with some doubts, for instance, the word has an uncommon meaning or is usually employed in different contexts/with different purposes. Scores less than $40\%$ mean  different, related, or unrelated semantics, which means that the word cannot be a candidate synonym in this context. It is worth noting that this fine-grained scoring schema emerged after different iterations of discussion with the linguists in order to create sound methodological guidelines to annotate the dataset with fuzzy values.

\begin{table}
\small
\centering
\begin{tabular}{| m{0.6cm}  m{ 5.8cm} | }
\hline 
\textbf{Score} & \textbf{Meaning} \\ \hline \hline 
100 & \textcolor{teal}{Same semantics, style, use}\\ \hline
 90 & \textcolor{teal}{Same semantics, style, less used}\\ \hline
 80 & \textcolor{teal}{Same semantics, style, rarely used}\\ \hline
 70 & \textcolor{teal}{Same semantics, style, not used}\\ \hline
 60 & \textcolor{orange}{Close semantics, weak style, uncommon}\\ \hline
 50 & \textcolor{orange}{Close semantics, not exact purpose} \\ \hline
 40 & \textcolor{purple}{Semantically related} \\ \hline
 30 & \textcolor{purple}{Semantically related (somehow)} \\ \hline
 20 & \textcolor{purple}{Semantically different} \\ \hline
 10 & \textcolor{purple}{Semantically very different} \\ \hline
  0 & {\small \textcolor{purple}{Semantically unrelated}} \\ \hline
\end{tabular} 
\caption{The fuzzy scoring scale - synonymy strength}
\label{table:SemanticMeanings}
\end{table}

\subsection{Linguists Agreement Evaluation}
\label{Section:LingAgreement}
The scoring of the 3K synonyms spanned over three months and took about 100 working hours for each linguist. The results of the four linguists are aggregated, and an average of all scores was computed.

To measure the (dis)agreements between linguists, we computed the Root Mean Squared Error (RMSE) and the Mean Average Error (MAE) between 
their scores (see table~\ref{table:RMSELiguists}). We also computed the RMSE and MAE between the scores of each linguist with the average score for the four linguists. Later, we will use the same model (i.e., the average of answers) as a baseline to evaluate our algorithm, (see subsection \ref{Section:AlgorithmEvaluationWithBaseline}).
The RMSE might be more commonly used than MAE in measuring the differences between scores, but we provide both metrics in this paper. The MAE scores treat differences equally, while RMSE penalizes large variations \cite{wang2018analysis}. 

\begin{table*}[h]
 \centering
 \scriptsize
    \begin{tabular}{|c|c|c|c|c|c|c|c|c|c|c|c|c|} \hline
     & \multicolumn{2}{|c}{\textbf{L1}} & \multicolumn{2}{|c}{\textbf{L2}} & \multicolumn{2}{|c}{\textbf{L3}} & \multicolumn{2}{|c}{\textbf{L4}} & \multicolumn{2}{|c}{\textbf{Avg}} &
    \multicolumn{2}{|c|}{\textbf{Algorithm}} \\
       \hline 
\multirow{-2}{*}{} & \multicolumn{1}{|c|}{\textbf{RMSE}} & \multicolumn{1}{|c}{\textbf{MAE}} & \multicolumn{1}{|c}{\textbf{RMSE}}  & \multicolumn{1}{|c}{\textbf{MAE}} & \multicolumn{1}{|c}{\textbf{RMSE}} & \multicolumn{1}{|c}{\textbf{MAE}} & \multicolumn{1}{|c}{\textbf{RMSE}} & \multicolumn{1}{|c}{\textbf{MAE}} &
 \multicolumn{1}{|c}{\textbf{RMSE}} & \multicolumn{1}{|c}{\textbf{MAE}} &
  \multicolumn{1}{|c}{\textbf{RMSE}} & \multicolumn{1}{|c|}{\textbf{MAE}} \\ \hline \hline
L1 & \multicolumn{1}{|l|}{\cellcolor[HTML]{efefef}} & \multicolumn{1}{|l|}{\cellcolor[HTML]{efefef}} & 0.19 & 0.14 & 0.19 & 0.14 & 0.22 & 0.16 & 0.13  & 0.10 & 0.35 & 0.30 \\ \hline
L2 & 0.19 & 0.14 & \multicolumn{1}{|l|}{\cellcolor[HTML]{efefef}} & \multicolumn{1}{|l|}{\cellcolor[HTML]{efefef}} & 0.16 & 0.12 & 0.20 & 0.15 & 0.10 & 0.11 & 0.31 & 0.26 \\ \hline
L3 & 0.19 & 0.14 & 0.16 & 0.12 & \multicolumn{1}{l}{\cellcolor[HTML]{efefef}} & \multicolumn{1}{|l|}{\cellcolor[HTML]{efefef}} & 0.20 & 0.16 & 0.11 & 0.08 & 0.32 & 0.26 \\ \hline
L4 & 0.22 & 0.16 & 0.20 & 0.15 & 0.20 & 0.15 & \multicolumn{1}{|l|}{\cellcolor[HTML]{efefef}} & \multicolumn{1}{|l|}{\cellcolor[HTML]{efefef}} & 0.13 & 0.08 & 0.39 & 0.34 \\ \hline
\multicolumn{1}{|c|}{\textbf{Avg}} & 0.13 & 0.10 & 0.10 & 0.08 & 0.11 & 0.08 & 0.13 & 0.11 & \multicolumn{1}{|l|}{\cellcolor[HTML]{efefef}} & \multicolumn{1}{|l|}{\cellcolor[HTML]{efefef}} & \textbf{0.32} & \textbf{0.27} \\ \hline

\multicolumn{1}{|c|}{\textbf{Algorithm}} & 0.35 & 0.30 & 0.31 & 0.26 & 0.32 & 0.26 & 0.39 & 0.34 & \textbf{0.32} & \textbf{0.27} & \multicolumn{1}{|l|}{\cellcolor[HTML]{efefef}} & \multicolumn{1}{|l|}{\cellcolor[HTML]{efefef}} \\ \hline 

\end{tabular}
 \caption{The Root Mean Squared Error (RMSE) and the Mean Average Error (MAE) between the scores of each linguist, the average scores of all linguists, and the scores of the algorithm.}
  \label{table:RMSELiguists}
\end{table*}

As shown in table~\ref{table:RMSELiguists}, linguists $L_2$ and $L_3$ have the closest RMSE to the average of all linguists. Linguists $L_1$ and $L_4$ have the highest RMSE distances if compared with the average scores. However, this does not indicate that they are more or less precise in their scores, it only shows that the scores of their answers deviate by the value stated by RMSE. Nevertheless, the RMSE of each linguist and the average ranges between $0.1$ and $0.13$. This indicates how much the scores of all linguists deviate from their average (i.e., which can be seen as an estimator of the standard deviation of errors between the linguist scores and the average of all linguists). 
It can be also noticed that the average deviation of the linguists and their average ranges between $0.31$ to $0.39$ from the algorithm. Though the algorithm deviates from the average score more than the individual linguists, the reported RMSE and MAE values are not considered high and further experiments are conducted to highlight if the difference between the scores is statistically significant.
 
To conform with this conclusion and to better understand the behavior of linguists in scoring these 3K synonyms, we perform a one-way ANOVA test (at $p<0.05$). This test determines if the difference between the linguists' scores is generated at random or if their scores are different consistently (i.e., significantly different). 

Post-hoc comparisons using the Tukey HSD test (using SPSS) indicated that the mean score for linguist $L_1$ (Mean = 0.4919, Standard Deviation = 0.34223) was significantly different than the other linguists (Mean = 0.4596, Standard Deviation = 0.31899). All included variables are following the normal distribution.

\section{Algorithm Overview}
\label{sec:algorithm}
The algorithm takes two inputs: a dictionary $D$, and a synset $S$. The output is a set of candidate synonyms $C$, each synonym $c_i$ is assigned a fuzzy value $f_i$. The dictionary $D$ itself is assumed to consist of set of synsets, $S_i$ $\in$ $D$. Each synset is a tuple $<t_1,..,t_n>$ of linguistic terms regardless of the language it belongs to. In this way, we can benefit from mono and multiple dictionaries and thesauri. In the first step, the algorithm extracts the candidate synonyms $C$, then it computes the fuzzy value $f_i$ for each synonym $c_i$. 

\subsection{Candidate Synonym Extraction}
For each term $t_i$ in synset $S$, the algorithm finds all cyclic paths at level $k$, where \textit{k} = 3, 4, 5,... n. That is, starting from $t_i$ as a root, a graph is constructed using $D$, at level $k$, and all paths starting and ending with $t_i$ are considered cyclic paths. If a term appears in any cyclic path, it is then considered a candidate synonym and is added to $C$.

\begin{figure*}[tb] 
\centering

 \makebox[\textwidth]{\includegraphics[width=0.85\paperwidth]{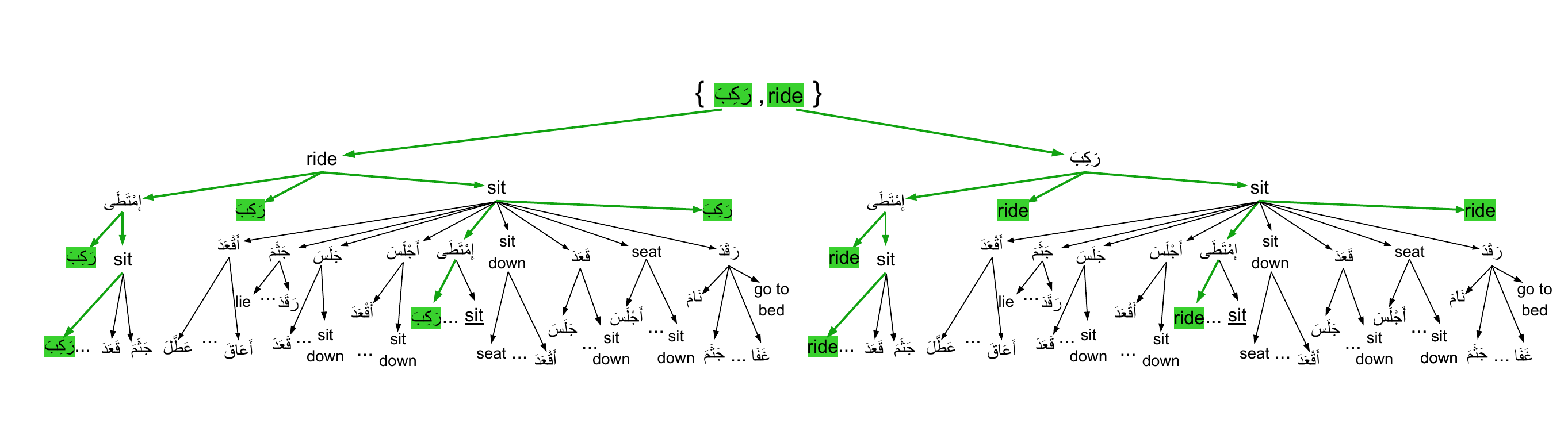}}
\caption{The cyclic paths for the \{{\scriptsize \<رَكِبَ>} , ride\} synset from AWN }
 \label{fig:ExampleGraph}
\end{figure*}

\textbf{Example:} Figure \ref{fig:ExampleGraph} illustrates the synset \{{\scriptsize \<رَكِبَ>}, ride\}, taken from the Arabic WordNet, and the generated graph at level 4 for each word in this synset. There are ten cyclic paths in this graph, highlighted as bold green lines, and shown below separately in Figure \ref{fig:WordSit}. The new terms participating in these ten cyclic paths are \{{\scriptsize\<إِمتَطَى>}, sit\}, which is the set $C$ of candidate synonyms.

\begin{figure}[h]
    \centering
\begin{center}
\fbox{\begin{varwidth}{\dimexpr\textwidth-2\fboxsep-2\fboxrule\relax}
{\scriptsize \<رَكِبَ>}→ride→sit→{\scriptsize \<رَكِبَ>} \\
{\scriptsize \<رَكِبَ>}→ride→sit→{\scriptsize \<إِمتَطَى>}→{\scriptsize \<رَكِبَ>} \\
{\scriptsize \<رَكِبَ>}→ride→{\scriptsize \<رَكِبَ>} \\
{\scriptsize \<رَكِبَ>}→ride→{\scriptsize \<إِمتَطَى>}→{\scriptsize \<رَكِبَ>} \\
{\scriptsize \<رَكِبَ>}→ride→{\scriptsize \<إِمتَطَى>}→sit→{\scriptsize \<رَكِبَ>} \\
ride→{\scriptsize \<رَكِبَ>}→sit→ride \\
ride→{\scriptsize \<رَكِبَ>}→sit→{\scriptsize \<إِمتَطَى>}→ride \\
ride→{\scriptsize \<رَكِبَ>}→ride \\
ride→{\scriptsize \<رَكِبَ>}→{\scriptsize \<إِمتَطَى>}→ride \\
ride→{\scriptsize \<رَكِبَ>}→{\scriptsize \<إِمتَطَى>}→sit→ride
\end{varwidth}}
\end{center}
    \caption{The ten cyclic paths extracted from the graph generated in Figure \ref{fig:ExampleGraph}.}
    \label{fig:WordSit}
\end{figure}

\subsection{Candidate Synonym Selection}
The intuition of our fuzzy model is that the more a candidate synonym appears in different cyclic 
paths and with different terms in $S$, the higher its fuzzy value, i.e., the stronger the synonymy. 
As such, to compute the fuzzy value $f_i$ for each $c_i$ in set $C$, we propose the following $Fuzzy$ 
function, which is based on two variables and two constant weights,  as in the following formula: 
\[ Fuzzy(f_i) = \theta_1 \cdot P_i  + \theta_2 \cdot Q_i  \]
where $P_i$ is the number of cyclic paths that $c_i$ appears in, divided by the total number of cyclic paths, and $Q_i$ is the number of root nodes $t$ that appear in the cyclic paths of $c_i$, divided by the total number of terms in the synset $S$.
$\theta_1$ and $\theta_2$ are two constant weights that we tuned using a 10-fold Cross-Validation (See section~\ref{sec:gridsearch}). 
The best values we found at level 3 and 4 are $(0.4, 0.6)$ and $(0.5, 0.5)$, respectively. As Figure~\ref{fig:ExampleGraph} illustrates, the term (sit) appears six times among the ten cyclic paths found at the level 4, and appears in two root 
nodes among the two synonyms in the original synset; 
and similarly for ({\scriptsize \<إِمتَطَى>}). Therefore, their fuzzy values are:

\begin{center}
\begin{varwidth}{\dimexpr\textwidth-2\fboxsep-2\fboxrule\relax}

$Fuzzy(sit) = \frac{6}{10} \times 0.5 + \frac{2}{2} \times 0.5 = 0.8$ \\ \\
$Fuzzy$({\scriptsize \<إِمتَطَى>}) 
$= \frac{6}{10} \times 0.5 + \frac{2}{2} \times 0.5 = 0.8$ 
\end{varwidth}
\end{center}

\subsection{Parameter Tuning}
\label{sec:gridsearch}
As our proposed $Fuzzy$ function depends on two constant weights ($\theta_1$ and $\theta_2$), our goal in this subsection is to find the best values of these $\theta_s$. The best values are those that enable the $Fuzzy$ function to produce fuzzy values as close to linguists' scores as possible. Thus, we used our dataset, which contains 3K candidate synonyms, each with a fuzzy value (i.e., the average of the four linguists). To generate a model with the best results, we varied the values of the parameters $\theta_1$ and $\theta_2$ by selecting their values within the range of $[0.1, 0.9]$ with a step of $0.1$ for each parameter. The total weight of both variables $\theta_1$ and $\theta_2$ should total to 1. This is because each of these variables is contributing to the score which ranges from 0 to 1. 

Table~\ref{table:Theta} shows the average RMSE and the average MAE values using a 10-fold Cross-Validation of the algorithm run on all combinations of the variables. The results show that the best combination is $0.5$ for $\theta_1$ and $0.5$ for $\theta_2$ which resulted in the lowest RMSE value of $0.32$, and the lowest MAE value of $0.27$ at level 4. For level 3, the best combination is $0.4$ for $\theta_1$ and $0.6$ for $\theta_2$ with value of $0.35$ for RMSE and $0.29$ for MAE. Thus, we complete the RMSE and MAE calculations with level 4, as the RMSE and MAE values in level 4 are better than in level 3. These are the weights that are used in the algorithm evaluations in the next section. 

\begin{table}[!h]
\begin{center}
\small

  \begin{tabularx}{0.4\textwidth} {| c | c | c | c |}
      \hhline{----}
     \multicolumn{2}{|c|}{$\theta_1 ,\theta_2 $ } & Level 4 &  Level 3 \\
    
     \hhline{----}
     \multirow{2}{4em}{ [0.1, 0.9]}&RMSE & 0.459 &0.377\\
     \hhline{~---}
      & MAE &0.375 &0.319\\

   \hhline{----}
     \multirow{2}{4em}{  [0.2, 0.8]}&RMSE & 0.408 &0.362\\
     \hhline{~---}
      & MAE &0.330 &0.304\\

 \hhline{----}
     \multirow{2}{4em}{ [0.3, 0.7]}&RMSE & 0.366 &0.352\\
     \hhline{~---}
      & MAE &0.299 &0.296\\
      
      \hhline{----}
     \multirow{2}{4em}{ [0.4, 0.6]}&RMSE & 0.336 & \textbf{0.349}\\
     \hhline{~---}
      & MAE &0.280 &\textbf{0.293}\\
      
       \hhline{----}
     \multirow{2}{4em}{ [0.5, 0.5]}&RMSE & \textbf{0.321} &0.352\\
     \hhline{~---}
      & MAE &\textbf{0.271}&0.296\\
      
       \hhline{----}
     \multirow{2}{4em}{ [0.6, 0.4]}&RMSE & 0.323 &0.363\\
     \hhline{~---}
      & MAE &0.271 &0.304\\
       \hhline{----}
     \multirow{2}{4em}{ [0.7, 0.3]}&RMSE & 0.343 &0.382\\
     \hhline{~---}
      & MAE &0.272 &0.316\\
      
       \hhline{----}
     \multirow{2}{4em}{ [0.8, 0.2]}&RMSE & 0.378 &0.407\\
     \hhline{~---}
      & MAE &0.302 &0.335\\
      
          \hhline{----}
     \multirow{2}{4em}{ [0.9, 0.1]}&RMSE & 0.425 &0.437\\
     \hhline{~---}
      & MAE & 0.335 & 0.357 \\
            \hhline{----}
 \end{tabularx}
 \end{center}
 \caption{Average RMSE and MAE with various values of $\theta_1$ and $\theta_2$ obtained using 10-fold Cross-Validation}
\label{table:Theta}
\end{table}

\section{Algorithm Evaluation}
\label{Section:AlgorithmEvaluation}
This section presents two experiments to evaluate the performance of our algorithm. The first experiment compares the results obtained by the algorithm with linguists' scores. The second experiment measures the accuracy of the
algorithm.
\subsection{Comparing the Algorithm with the Baseline}
\label{Section:AlgorithmEvaluationWithBaseline}
This experiment compares the results of our algorithm with the average of the linguists' scores (as a baseline) that we presented in section \ref{Section:LingAgreement}. 

Table~\ref{table:RMSELiguists} shows $0.32$ RMSE and $0.27$ MAE scores of the algorithm against the linguists' average. To understand the algorithm's $0.32$ RMSE, one can notice that the RMSE difference between $L_2$ and $L_4$ is $0.20$, and between $L_1$ and $L_4$ is $0.22$. The RMSE difference between each pair of linguists ranges from $0.16$ to $0.22$. Now, the RMSE difference between the algorithm and the average of the linguists is $0.32$. This means that the algorithm has only $0.10$ more difference if compared with the RMSE variation between linguists. 

Similarly, to understand the $0.27$ MAE between the algorithm and the linguists, one can notice that the MAE between the four linguists themselves ranges from $0.12$ to $0.16$. Both RMSE and MAE, confirm the variation between the algorithm and the average of linguists. This illustrates that the algorithm's scores are close to the linguists' scores.

Nevertheless, as noted in section \ref{Section:LingAgreement}, the variations between linguists' scores, as well as the algorithm, do not tell us whether a linguist is better or more accurate than the others, which is because synonymy is a subjective notion. However, being close to the linguists' variations is a good indication that the algorithm scores are realistic. Next, we compare the behavior of the algorithm with the linguists' behavior in scoring synonyms, which provides an additional evaluation.

\textbf{Testing the algorithm's \textit{behavior}}: to further understand the algorithm's behavior, we need to test whether the scores of the algorithm are statistically significant, i.e., the scores were consistent or resulted at random. In other words, we need to test whether the algorithm is consistently giving scores and behaving like a linguist - regardless of the differences in RMSE and MAE.

We performed a one-way ANOVA test (at $p<0.05$) to check if there is a statistical difference between the algorithm and the other linguists. Before conducting this test, we first needed to check if all the linguists' and the algorithm's scores follow a normal distribution, or if there are no outliers, which are the main assumptions to conduct a one-way ANOVA test. Our result of the normality test (using SPSS) indicated that the scores of the algorithm are not normally distributed. Thus, we performed a univariate and multivariate outlier analysis. The results (using SPSS) indicated that there are no outliers, which means that the non-normality of the algorithm's scores are due to skewness in the data and not because of outliers. Therefore, the one-way ANOVA test can be applied, as explained by \citet{tabachnick2001using}: \textit{``it is assumed that the data has a normal distribution, however, note that violations of the normality assumption are not fatal and the result of the significant test is still reliable as long as non-normality is caused by skewness and not outliers''}. 

The post-hoc comparisons (using the Tukey HSD test, in SPSS) indicated that the mean score for the algorithm (Mean = 0.4535, Standard Deviation = 0.16416) was significantly different only with linguist $L_1$ (Mean = 0.4919, Standard Deviation = 0.34223). This indeed confirms the findings shown in the previous section in which linguist $L_1$ has significantly different scores than the other linguists. In other words, the algorithm has shown to be not statistically different with the other linguists and their average (i.e., the baseline). Being not statistically different means that the algorithm's behavior in scoring synonyms is similar to the behavior of the linguists, except for linguist $L_1$.

To sum up, the variation between the scores of the algorithm and the linguists (using RMSE and MAE) are close to those between the linguists themselves. The one-way ANOVA test also confirms that the algorithm behaves as a linguist.

\subsection{Accuracy Evaluation}
We measure the accuracy of the algorithm in terms of retrieved words for each synset, by masking a synonym in a given synset, then try predicting it again. Masking is the process of removing a synonym from a synset, and then measure whether the masked term is retrieved back. The accuracy of the algorithm is determined by the rank of the masked term. Ideally, if every masked term is retrieved with the highest (i.e., top) rank, it means the accuracy is $100\%$.

\subsubsection{Experiment Setup}
We used the 10K synsets in the Arabic WordNet (AWN), and we conducted four masking experiments. For every synset in the 10K AWN's synsets, we calculated the frequency of each synonym (Arabic and English), then selected the synonyms with (highest, lowest, average, and random) frequencies in each synset to conduct the experiment. The frequency of a term is the number of synsets in which this term appears. We considered synsets that contain more than two synonyms, regardless of the language. That is, the experiment was conducted on both Arabic and English terms. Terms with the frequency of 1 (i.e., appeared in one synset only) are not selected. The number of synsets that are longer than two terms, and with a term with a frequency  more than 1 are $7,219$, while the number of synsets longer than two terms, and with a term with lowest frequency are only $1,085$. Similarly, we selected synsets with average and random term frequencies, $5,207$ and $4,153$, respectively. Table~\ref{table:SectionResultsMAX} shows the results of the masking experiments.

The algorithm was applied individually for each synset by eliminating (i.e., masking) a term, in this synset, and retrieving back the top-ranked term using the algorithm. That is, given a term $c_{1}$ in synset $s_{n}$, $c_{1}$ will be eliminated from $s_{n}$, then we compute the fuzzy value of $c_{1}$ using our algorithm and check if the algorithm was able to retrieve it with highest fuzzy value (i.e., top rank) among other possible candidate synonyms for $s_{n}$. In this way, the algorithm is applied on synsets by masking terms with highest, lowest, average, and random frequencies, at level 3; and repeated at level 4, as shown in Table~\ref{table:SectionResultsMAX}.

\subsubsection{Results}

The accuracy of the algorithm was calculated as a ratio of the correctly retrieved synonyms (i.e., top rank) from all samples. For example, the algorithm was able to retrieve 7,157 (99.1\%) of the masked terms with highest frequencies at level 3 with the top ranking (i.e., highest fuzzy values).

The results in Table~\ref{table:SectionResultsMAX} illustrate that the lower the frequency of a term in the lexicon the lower the accuracy, which is because the connectivity of less frequent terms yields less fuzzy values by the algorithm. This does not mean that the masked terms were not retrieved by the algorithm, but rather, they are not ranked as the top (highest fuzzy values). The accuracy at level 4 decreases because the synonymy graph at this level becomes larger, and thus it contains more candidate synonyms.

It is important to remark that the algorithm was able to obtain high accuracy in this experiment but the accuracy evaluation heavily depends on the structure of the used lexicon, which is AWN in our case. Changing the dictionary, by adding more synonymy/translation relations yields to constructing a different graph, thus different accuracy is expected.


\begin{table}[!h]
    \centering
    \small
    \begin{tabular}{|l|c|c|c|c|}
    \hline
         \makecell[c]{Experiment} &
         \makecell[c]{Sample \\ Size} & 
         \makecell[c]{Accuracy \\ at Level 3}  & 
         \makecell[c]{Accuracy \\ at Level 4}  \\ \hline\hline
         Exp.1 (Highest) & $7,219$ & $99.1\%$ &  $95.2\%$ \\ \hline
         Exp.2 (Average) & $5,207$ &  $98.7\%$ &  $92.0\%$ \\ \hline
         Exp.3 (Lowest) & $1,085$ & $88.4\%$ &  $62.0\%$ \\ \hline
         Exp.4 (Random) & $4,153$ & $98.1\%$ &  $89.3\%$ \\ \hline
    \end{tabular}
 \caption{The accuracy of the algorithm using the masking experiment with the highest, average, lowest, and random frequencies within each synset.}
 \label{table:SectionResultsMAX}
\end{table}


\section{Conclusion}


We presented a benchmark dataset and an algorithm to extract synonyms and fuzzy values. The benchmark dataset consists of 3K candidate synonyms for 500 synsets, each candidate synonym was annotated with a fuzzy value by four linguists. The dataset is important for measuring how much linguists disagree on synonymy, which ranged between $0.16-0.22$ for RMSE and $0.12-0.16$ for MAE. These measures were also used as a baseline to evaluate our algorithm. 
The algorithm presented in this paper aims to enrich a given mono/multilingual synset with more synonyms. 
Our evaluation shows that our algorithm behaves as linguists in producing fuzzy values, and the fuzzy scores are also close to those of the linguists. The accuracy evaluation illustrates that it is highly accurate.

\section{Limitations and Future Work}
The current version of our algorithm neglects the effect of diacritics in the Arabic language \cite{JZAA18}, so that a word with different diacritics is considered as different, like {\scriptsize \< كَتب>}, {\scriptsize \< كتَب>}, even if they are the same. Thus, we plan to enhance the algorithm to consider the characteristics of the Arabic language, and consider synonyms in MSA and Arabic dialects as described in \cite{,EJHZ22,JHRAZ17,JZHNW22}. 

\section*{Acknowledgment}
\label{sec:ack}
We acknowledge the support of the Research Committee at Birzeit University (No. 2021/49), and would like to thank Taymaa Hammouda and Muhannad Yaseen for the technical and statistical support, and all students who helped in the annotation process, especially Tamara Qaimari, Asala Hamed, Ahd Muhtasib, Doa Shwiki, Shaimaa Hamayel, Hiba Zayed, Rwaa Zaid, and others.

\bibliography{custom}

\begin{thebibliography}{34}
\expandafter\ifx\csname natexlab\endcsname\relax\def\natexlab#1{#1}\fi

\bibitem[{Al-Hajj and Jarrar(2021)}]{HJ21b}
Moustafa Al-Hajj and Mustafa Jarrar. 2021.
\newblock \href {https://doi.org/10.26615/978-954-452-072-4_005}
  {Arabglossbert: Fine-tuning bert on context-gloss pairs for wsd.}
\newblock In \emph{Proceedings of the International Conference on Recent
  Advances in Natural Language Processing (RANLP 2021)}, pages 40--48, Online.
  INCOMA Ltd.

\bibitem[{Al-Matham and Al-Khalifa(2021)}]{al2021synoextractor}
Rawan~N Al-Matham and Hend~S Al-Khalifa. 2021.
\newblock Synoextractor: a novel pipeline for arabic synonym extraction using
  word2vec word embeddings.
\newblock \emph{Complexity}, 2021.

\bibitem[{Alizadeh{-}Q et~al.(2021)Alizadeh{-}Q, Minaei{-}Bidgoli, Hossayni,
  Akbarzadeh{-}T., Recupero, Rajati, and Gangemi}]{iran2021}
Yousef Alizadeh{-}Q, Behrouz Minaei{-}Bidgoli, Sayyed{-}Ali Hossayni,
  Mohammad{-}R. Akbarzadeh{-}T., Diego~Reforgiato Recupero, Mohammad~Reza
  Rajati, and Aldo Gangemi. 2021.
\newblock \href {http://arxiv.org/abs/2104.10660} {Interval probabilistic fuzzy
  wordnet}.
\newblock \emph{CoRR}, abs/2104.10660.

\bibitem[{Darwish et~al.(2021)Darwish, Habash, Abbas, Al-Khalifa, Al-Natsheh,
  Bouamor, Bouzoubaa, Cavalli-Sforza, El-Beltagy, El-Hajj, Jarrar, and
  Mubarak}]{DH21}
Kareem Darwish, Nizar Habash, Mourad Abbas, Hend Al-Khalifa, Huseein~T.
  Al-Natsheh, Houda Bouamor, Karim Bouzoubaa, Violetta Cavalli-Sforza,
  Samhaa~R. El-Beltagy, Wassim El-Hajj, Mustafa Jarrar, and Hamdy Mubarak.
  2021.
\newblock \href {https://doi.org/10.1145/3447735} {A panoramic survey of
  natural language processing in the arab worlds}.
\newblock \emph{Commun. ACM}, 64(4):72–81.

\bibitem[{Elkateb et~al.(2006)Elkateb, Black, Vossen, Farwell, Rodr{\'\i}guez,
  Pease, and Alkhalifa}]{ArabicWordNet}
Sabry Elkateb, William Black, Piek Vossen, David Farwell, H~Rodr{\'\i}guez,
  A~Pease, and M~Alkhalifa. 2006.
\newblock Arabic wordnet and the challenges of arabic.
\newblock In \emph{Proceedings of Arabic NLP/MT Conference, London, UK}, pages
  665--670.

\bibitem[{Ercan and Haziyev(2019)}]{ercan2019synset}
Gonenc Ercan and Farid Haziyev. 2019.
\newblock Synset expansion on translation graph for automatic wordnet
  construction.
\newblock \emph{Information Processing \& Management}, 56(1):130--150.

\bibitem[{Flati and Navigli(2012)}]{flati2012cqc}
Tiziano Flati and Roberto Navigli. 2012.
\newblock The cqc algorithm: Cycling in graphs to semantically enrich and
  enhance a bilingual dictionary.
\newblock \emph{Journal of Artificial Intelligence Research}, 43:135--171.

\bibitem[{Haff et~al.(2022)Haff, Jarrar, Hammouda, and Zaraket}]{EJHZ22}
Karim~El Haff, Mustafa Jarrar, Tymaa Hammouda, and Fadi Zaraket. 2022.
\newblock Curras + baladi: Towards a levantine corpus.
\newblock In \emph{Proceedings of the International Conference on Language
  Resources and Evaluation (LREC 2022)}, Marseille, France.

\bibitem[{Helou et~al.(2016)Helou, Palmonari, and Jarrar}]{APJ16}
Mamoun~Abu Helou, Matteo Palmonari, and Mustafa Jarrar. 2016.
\newblock \href {https://doi.org/10.1613/jair.4789} {Effectiveness of automatic
  translations for cross-lingual ontology mapping}.
\newblock \emph{Journal of Artificial Intelligence Research}, 55(1):165--208.

\bibitem[{Helou et~al.(2014)Helou, Palmonari, Jarrar, and Fellbaum}]{HPJF14}
Mamoun~Abu Helou, Matteo Palmonari, Mustafa Jarrar, and Christiane Fellbaum.
  2014.
\newblock \href
  {https://www.researchgate.net/publication/288120715_Towards_building_lexical_ontology_via_cross-language_matching}
  {Towards building lexical ontology via cross-language matching}.
\newblock In \emph{Proceedings of the 7th Conference on Global WordNet}, pages
  346--354. Global WordNet Association.

\bibitem[{Hossayni et~al.(2020)Hossayni, Akbarzadeh{-}T., Recupero, Gangemi,
  del Acebo, and de~la Rosa~i Esteva}]{iran2020}
Sayyed{-}Ali Hossayni, Mohammad{-}R. Akbarzadeh{-}T., Diego~Reforgiato
  Recupero, Aldo Gangemi, Esteve del Acebo, and Josep~Llu{\'{\i}}s de~la Rosa~i
  Esteva. 2020.
\newblock \href {http://arxiv.org/abs/2006.04042} {An algorithm for
  fuzzification of wordnets, supported by a mathematical proof}.
\newblock \emph{CoRR}, abs/2006.04042.

\bibitem[{Jarrar(2005)}]{J05}
Mustafa Jarrar. 2005.
\newblock \href {http://www.jarrar.info/phd-thesis/JarrarPhDThesisV167.pdf}
  {\emph{Towards Methodological Principles for Ontology Engineering}}.
\newblock Ph.D. thesis, Vrije Universiteit Brussel.

\bibitem[{Jarrar(2011)}]{J11}
Mustafa Jarrar. 2011.
\newblock \href {http://www.jarrar.info/publications/J11.pdf} {Building a
  formal arabic ontology (invited paper)}.
\newblock In \emph{Proceedings of the Experts Meeting on Arabic Ontologies and
  Semantic Networks}. ALECSO, Arab League.

\bibitem[{Jarrar(2020)}]{J20}
Mustafa Jarrar. 2020.
\newblock \href
  {https://www.researchgate.net/publication/351335422_Digitization_of_Arabic_Lexicons}
  {\emph{Digitization of Arabic Lexicons}}, pages 214--217. UAE Ministry of
  Culture and Youth.

\bibitem[{Jarrar(2021)}]{J21}
Mustafa Jarrar. 2021.
\newblock \href {https://doi.org/10.3233/AO-200241} {The arabic ontology - an
  arabic wordnet with ontologically clean content}.
\newblock \emph{Applied Ontology Journal}, 16(1):1--26.

\bibitem[{Jarrar and Amayreh(2019)}]{JA19}
Mustafa Jarrar and Hamzeh Amayreh. 2019.
\newblock \href {https://doi.org/10.1007/978-3-030-23281-8_19} {An
  arabic-multilingual database with a lexicographic search engine}.
\newblock In \emph{The 24th International Conference on Applications of Natural
  Language to Information Systems (NLDB 2019)}, volume 11608 of \emph{LNCS},
  pages 234--246. Springer.

\bibitem[{Jarrar et~al.(2019)Jarrar, Amayreh, and McCrae}]{JAM19}
Mustafa Jarrar, Hamzeh Amayreh, and John~P. McCrae. 2019.
\newblock \href {http://www.jarrar.info/publications/JAM19.pdf} {Representing
  arabic lexicons in lemon - a preliminary study}.
\newblock In \emph{The 2nd Conference on Language, Data and Knowledge (LDK
  2019)}, volume 2402, pages 29--33. CEUR Workshop Proceedings.

\bibitem[{Jarrar et~al.(2017)Jarrar, Habash, Alrimawi, Akra, and
  Zalmout}]{JHRAZ17}
Mustafa Jarrar, Nizar Habash, Faeq Alrimawi, Diyam Akra, and Nasser Zalmout.
  2017.
\newblock \href {https://doi.org/10.1007/S10579-016-9370-7} {Curras: An
  annotated corpus for the palestinian arabic dialect}.
\newblock \emph{Journal Language Resources and Evaluation}, 51(3):745--775.

\bibitem[{Jarrar et~al.(2021)Jarrar, Karajah, Khalifa, and Shaalan}]{JKKS21}
Mustafa Jarrar, Eman Karajah, Muhammad Khalifa, and Khaled Shaalan. 2021.
\newblock \href {http://www.jarrar.info/publications/JKKS21.pdf} {Extracting
  synonyms from bilingual dictionaries}.
\newblock In \emph{Proceedings of the 11th International Global Wordnet
  Conference (GWC2021)}, pages 215--222. Global Wordnet Association.

\bibitem[{Jarrar et~al.(2018)Jarrar, Zaraket, Asia, and Amayreh}]{JZAA18}
Mustafa Jarrar, Fadi Zaraket, Rami Asia, and Hamzeh Amayreh. 2018.
\newblock \href {https://doi.org/10.1145/3242177} {Diacritic-based matching of
  arabic words}.
\newblock \emph{ACM Asian and Low-Resource Language Information Processing},
  18(2):10:1--10:21.

\bibitem[{Jarrar et~al.(2022)Jarrar, Zaraket, Hammouda, Alavi, and
  Waahlisch}]{JZHNW22}
Mustafa Jarrar, Fadi~A Zaraket, Tymaa Hammouda, Daanish~Masood Alavi, and
  Martin Waahlisch. 2022.
\newblock \href {https://doi.org/10.48550/ARXIV.2212.06468} {Lisan: Yemeni,
  irqi, libyan, and sudanese arabic dialect copora with morphological
  annotations}.

\bibitem[{Khodak et~al.(2017)Khodak, Risteski, Fellbaum, and
  Arora}]{khodak2017automated}
Mikhail Khodak, Andrej Risteski, Christiane Fellbaum, and Sanjeev Arora. 2017.
\newblock Automated wordnet construction using word embeddings.
\newblock In \emph{Proceedings of the 1st Workshop on Sense, Concept and Entity
  Representations and their Applications}, pages 12--23.

\bibitem[{Maru et~al.(2022)Maru, Conia, Bevilacqua, and Navigli}]{nibbling2022}
Marco Maru, Simone Conia, Michele Bevilacqua, and Roberto Navigli. 2022.
\newblock \href {https://doi.org/10.18653/v1/2022.acl-long.324} {{N}ibbling at
  the hard core of {W}ord {S}ense {D}isambiguation}.
\newblock In \emph{Proceedings of the ACL2022 (Vol.1)}, pages 4724--4737,
  Dublin, Ireland. ACL.

\bibitem[{Miller et~al.(1990)Miller, Beckwith, Fellbaum, Gross, and
  Miller}]{miller1990introduction}
George~A Miller, Richard Beckwith, Christiane Fellbaum, Derek Gross, and
  Katherine~J Miller. 1990.
\newblock Introduction to wordnet: An on-line lexical database.
\newblock \emph{International journal of lexicography}, 3(4):235--244.

\bibitem[{Mohammed(2020)}]{Mohammed2020}
Nora Mohammed. 2020.
\newblock \href {https://doi.org/10.34028/iajit/17/1/6} {Extracting word
  synonyms from text using neural approaches}.
\newblock \emph{International Arab Journal of Information Technology}, 17.

\bibitem[{Naser-Karajah et~al.(2021)Naser-Karajah, Arman, and Jarrar}]{KAJ21}
Eman Naser-Karajah, Nabil Arman, and Mustafa Jarrar. 2021.
\newblock \href {https://doi.org/10.1109/ICIT52682.2021.9491713} {Current
  trends and approaches in synonyms extraction: Potential adaptation to
  arabic}.
\newblock In \emph{Proceedings of the 2021 International Conference on
  Information Technology (ICIT)}, pages 428--434, Amman, Jordan. IEEE.

\bibitem[{Oliveira and Gomes(2014)}]{Oliveira14}
Hugo~Gonçalo Oliveira and Paulo Gomes. 2014.
\newblock \href {https://doi.org/10.1007/s10579-013-9249-9} {Eco and onto.pt: A
  flexible approach for creating a portuguese wordnet automatically}.
\newblock \emph{Language Resources and Evaluation}, 48.

\bibitem[{Oliveira and Santos(2016)}]{Oliveira2016}
Hugo~Gonçalo Oliveira and Fábio Santos. 2016.
\newblock Discovering fuzzy synsets from the redundancy in different
  lexical-semantic resources.
\newblock In \emph{Proceedings of LREC 2016}, Paris, France. ELRA.

\bibitem[{Tabachnick and Fidell(2001)}]{tabachnick2001using}
Barbara~G Tabachnick and LS~Fidell. 2001.
\newblock Using multivariate statististics.
\newblock \emph{Allyn \&Bacon A Pearson Education Company: Boston}.

\bibitem[{Tarouti and Kalita(2016)}]{tarouti16}
Feras~Al Tarouti and Jugal Kalita. 2016.
\newblock \href {https://doi.org/10.18653/v1/w16-1204} {Enhancing automatic
  wordnet construction using word embeddings}.

\bibitem[{Torregrosa et~al.(2019)Torregrosa, Arcan, Ahmadi, and
  McCrae}]{torregrosa2019tiad}
Daniel Torregrosa, Mihael Arcan, Sina Ahmadi, and John~P McCrae. 2019.
\newblock Tiad 2019 shared task: Leveraging knowledge graphs with neural
  machine translation for automatic multilingual dictionary generation.
\newblock \emph{Translation Inference Across Dictionaries}.

\bibitem[{Villegas et~al.(2016)Villegas, Melero, Bel, and
  Gracia}]{villegas2016leveraging}
Marta Villegas, Maite Melero, N{\'u}ria Bel, and Jorge Gracia. 2016.
\newblock Leveraging rdf graphs for crossing multiple bilingual dictionaries.
\newblock In \emph{Proceedings of LREC2016}, pages 868--876.

\bibitem[{Wang and Lu(2018)}]{wang2018analysis}
Weijie Wang and Yanmin Lu. 2018.
\newblock Analysis of the mean absolute error (mae) and the root mean square
  error (rmse) in assessing rounding model.
\newblock In \emph{IOP conference series: materials science and engineering},
  volume 324. IOP Publishing.

\bibitem[{Wu and Zhou(2003)}]{wu2003optimizing}
Hua Wu and Ming Zhou. 2003.
\newblock Optimizing synonym extraction using monolingual and bilingual
  resources.
\newblock In \emph{Proceedings of the second international workshop on
  Paraphrasing}, pages 72--79.

\end{thebibliography}
\bibliographystyle{acl_natbib}




\end{document}